\documentclass{article}

\usepackage[final,nonatbib]{nips_2018}

\usepackage[utf8]{inputenc} % allow utf-8 input
\usepackage[T1]{fontenc}    % use 8-bit T1 fontshttps://www.overleaf.com/project/5bae91cc8c64aa2923b47e89
\usepackage{hyperref}       % hyperlinks
\usepackage{url}            % simple URL typesetting
\usepackage{booktabs}       % professional-quality tables
\usepackage{amsfonts}       % blackboard math symbols
\usepackage{nicefrac}       % compact symbols for 1/2, etc.
\usepackage{microtype}      % microtypography
\usepackage{graphicx}
\usepackage{amssymb}
\usepackage{amsmath}
\usepackage{algorithm}
\usepackage{algpseudocode}
\usepackage{tikz}
\usetikzlibrary{bayesnet}
\usepackage{wrapfig}
\usepackage{graphicx}
\usepackage{caption}
\usepackage{subcaption}

\title{Efficient transfer learning and online adaptation with latent variable models for continuous control}

\author{
  Christian F. Perez \\
  Uber AI Labs\\
  San Francisco, CA 94105 \\
  \texttt{cfp@uber.com} \\
  \And
   Felipe Petroski Such \\
     Uber AI Labs\\
  San Francisco, CA 94105 \\
   \texttt{felipe.such@uber.com}
   \AND
   Theofanis Karaletsos \\
     Uber AI Labs\\
  San Francisco, CA 94105 \\
   \texttt{theofanis@uber.com}
}

\newcommand{\s}{\mathbf s_t}
\newcommand{\ns}{\mathbf s_{t+1}}
\newcommand{\action}{\mathbf a_t}
\newcommand{\env}{\mathbf e_k}

\begin{document}

\maketitle

\begin{abstract}

Traditional model-based RL relies on hand-specified or learned models of transition dynamics of the environment. These methods are sample efficient and facilitate learning in the real world but fail to generalize to subtle variations in the underlying dynamics, e.g., due to differences in mass, friction, or actuators across robotic agents or across time. We propose using variational inference to learn an explicit latent representation of unknown environment properties that accelerates learning and facilitates generalization on novel environments at test time. We use Online Bayesian Inference of these learned latents to rapidly adapt online to changes in environments without retaining large replay buffers of recent data. Combined with a neural network ensemble that models dynamics and captures uncertainty over dynamics, our approach demonstrates positive transfer during training and online adaptation on the continuous control task HalfCheetah.
\end{abstract}

\section{Introduction}
The ideal reinforcement learning algorithm learns efficiently with little data, generalizes well to new environments, and readily adapts to changing conditions. Model-free methods have achieved impressive results, but may require millions of observations during training \cite{Mnih2015}. Model-based methods are sample efficient, but often perform worse than model-free policies \cite{Deisenroth2011, Deisenroth2011a, Gal2016}. Both methods can suffer from over-fitting to training conditions, yielding agents that perform poorly when test conditions differ. If model-based agents can be made more adaptable to changing conditions, we can benefit from  sample-efficient learning across more realistic and dynamic environmental conditions.

We explore learning across environments in which transition dynamics can be expected to vary in systematic ways. The rules of physics do not change, but unknown physical parameters (e.g., friction, mass, actuator gain) can easily vary across agents or across time. Instead of requiring a human to specify how environment conditions affect dynamics, we would like an algorithm to learn these relationships and pool information for faster learning and more robust control. Hierarchical probabilistic models facilitate incorporation of prior knowledge about such information and can help transfer knowledge between tasks (for an early example see \cite{haruno2001mosaic} and more recently \cite{killian2017robust, doshi2016hidden, hausman2018}.) We propose hierarchical probabilistic model-based control with dynamics models that (1) use auxiliary latent variables to learn more efficiently by transferring learned dynamics across environments, and (2) generalize to novel environments through inference of the latent variables. In our model, the neural network is shared across all instances/variations of the environment, and private latent variables capture instance-specific variability.
To handle dynamics changing over time, we employ online Bayesian inference to equip the agent with the ability to infer changing environments on the fly. The resulting high-performing adaptive agents learn efficiently within tens of episodes across related continuous control environments and generalize to novel and changing dynamics.

In contrast to previous work on latent variable MDPs with small discrete action spaces \cite{doshi2016hidden, killian2017robust, yao2018}, our controllers achieve high reward on challenging robotic environments with continuous control tasks and are more sample-efficient than \cite{hausman2018}. As a pragmatic alternative to Bayesian neural networks, \emph{deep ensembles} \cite{balaji2017} are a promising approach that can address \emph{model bias} and uncertainty quantification in model-based methods, which can close the gap with model-free performance \cite{Chua2018, Gal2016, nagabandi2018}. Furthermore, we leverage variational inference to quickly learn the hidden latent environment variables within a single episode. This allows test-time adaptation to dynamically changing conditions, similar to \cite{Clavera2018}.

\section{Dynamics models with auxiliary latent variables}

\begin{wrapfigure}{r}{4cm}
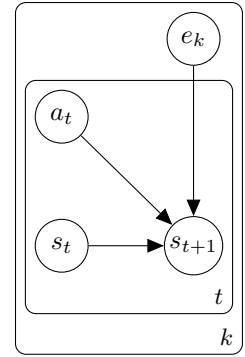

\centering
    \tikz{
        % Nodes
        \node[latent] (s)   {$s_t$};
        \node[latent, above=of s]   (a)   {$a_t$};
        \node[latent, right=of s]   (ns) {$s_{t+1}$};
        \node[latent, above=2 of ns] (e) {$e_k$};

       \edge {s,a,e} {ns};

       \plate {mdp} {(s)(ns)(a)} {$t$};
       \plate {env} {(e)(mdp)} {$k$}
    }
    \caption{Graphical model of multiple environment transition dynamics probability model.}
    \label{fig:model}
\end{wrapfigure}

For environments with different dynamics, we define $\env \in \mathbb{R}^{d_e}$ as a latent variable representation of the relevant degrees of freedom in the dynamics. In general, we do not have access to the parameters of the transition function, and so treat them as hidden latent variables.

For an agent acting in an unknown/new environment (specifically with an unknown transition function), a robust dynamics model explicitly accounts for beliefs about the environment, actor properties and environment-conditional dynamics and can marginalize over them:
\begin{equation}
p( \mathbf{s}_{0:T+1}, \mathbf{a}_{0:T},\env) = p(\env) p(\mathbf s_0) \prod \limits_{t=0}^{T} p(\ns|\s,\ \action, \env) p(\action|\s,\env)
\label{eq:dynamics}
\end{equation}
where $p(\env) = \mathcal{N}(0,I)$ and the initial state distribution $p(\mathbf s_0)$ is given by the environment. Recent work has shown neural network ensembles capture uncertainty and avoid overfitting in supervised \cite{balaji2017} and reinforcement learning problems \cite{Chua2018}. Given their straightforward implementation and reliable training protocols, we model $p(\ns |\s, \action, \env)$ as an ensemble of probabilistic networks incorporating variational inference as an alternative to a fully Bayesian neural network (see Sec.~\ref{app:ensembles}).

As the agent acts, new transitions $\mathcal{D}^*$ are collected and added to the dataset $\mathcal{D} = \mathcal{D} \cup \mathcal{D}^*$. For each transition in a new environment, we update the posterior over $\env$ via Bayes' rule,

\begin{eqnarray}
p(e_k|\mathcal{D}) &=& \frac{p(\mathcal{D}|e_k) p(e_k)}{p(\mathcal{D})} = \frac{p(\mathcal{D}|e_k) p(e_k)}{\int p(\mathcal{D}|e_k)  p(e_k) d e_k}
\label{eq:posterior}
\end{eqnarray}

In general, the marginal data likelihood $p(\mathcal{D})$ involves an intractable integral, but the posterior (Eq.~\ref{eq:posterior}) can be approximated by stochastic variational inference ({\bf SVI}).

% \subsection{Online Bayesian Learning}
To avoid retaining a large buffer of past experience $\mathcal{D}$, we utilize Online Bayesian Inference: We posit the posterior inference problem in sequential fashion for datasets $\mathcal{D}_{t}$ arriving successively, where the posterior at time $t$ becomes the prior for time $t+1$:
\begin{eqnarray}
p(e_k|\mathcal{D}_{t+1}) = \frac{p(\mathcal{D}_{t+1}|e_k) p(e_k|\mathcal{D}_{t})}{p(\mathcal{D}_{t+1})}
\end{eqnarray}

Unfortunately, in this setting we still have to keep $\mathcal{D}_t$ around to represent the posterior. However, using approximate inference techniques explained in the next section we can represent that partial posterior in a parametric form, allowing us to {\it forget} data from the past when learning about the current environment.

\subsection{Bayesian learning with variational inference}

% The generative model in Figure \ref{fig:model} describes an partially observable MDP where the hidden latent variable $e_k$ is sampled once per episode. Specifically, during training
% \begin{enumerate}
%     \item Sample an environment $k$ from a finite set.
%     \item Sample from the prior $e_k \sim p(e_k)$
%     \item Sample a trajectory (episode) $s_t, a_t, r_t \sim P(\tau|e_k)$ using the current dynamics model
%     \item
% \end{enumerate}

Formally, transitions are sampled iid from the intractable distribution $p(\env|\mathcal{D})$ that we approximate with $q_\phi(\env)$ parameterized by a multivariate normal where $\mathbf{\phi}=\{\mu_q, \Sigma_q\}$. The learning objective is to maximize the marginal likelihood of observed transitions with respect to $\theta$ and $\phi$. We can maximize the evidence lower bound ({\bf ELBO}) to this,
% \begin{equation}
\begin{align}
\begin{split}
    \log p(\mathcal D) &= \sum_{t=0}^T \log p(\ns|\s,\action)\\
    &\geq \mathbb{E}_{\env \sim q_\phi(\env)}\left[ \sum_{t=0}^{T}  \log p_\theta(\ns|\s,\action,\env) \right] - \mathrm{KL}\big(q_\phi(\env) || p(\env)\big).
\label{eq:objective}
\end{split}
\end{align}
 For simplicity, we choose the prior $p(\env)$ and variational distribution $q_\phi(\env)$ to be Gaussian with diagonal covariance. During online learning, the prior is updated from new data $\mathcal D_t$ and we optimize a subtly different objective holding network parameters $\theta$ fixed to find an approximate posterior:
\begin{equation}
\begin{gathered}
p(\env|\mathcal D_{t+1}) \ge q_{\phi^*}(\env) \\
    \mathrm{s.t.} \, \phi^* = \mathrm{argmax}_{\phi} \mathbb{E}_{\env \sim q_\phi(\env)}\left[ \log p_\theta(\ns|\s,\action,\env) \right] - \mathrm{KL}\big(q_\phi(\env |\mathcal{D}_{t+1}) || p(\env|\mathcal{D}_{t})\big).
\label{eq:objective_online}
\end{gathered}
\end{equation}

%  \begin{equation}
%     p(\env|\mathcal D_{t+1}) \approx \mathrm{argmax}_{q_\phi} \mathbb{E}_{\env \sim q_\phi(\env)}\left[ \log p_\theta(\ns|\s,\action,\env) \right] - \mathrm{KL}\big(q_\phi(\env |\mathcal{D}_{t+1}) || p(\env|\mathcal{D}_{t})\big).
% \end{equation}

\subsection{Control with Auxiliary Variable Models}

Given a learned dynamics model, agents can plan into the future by recursively predicting future states $\mathbf s_{t+1}, ..., \mathbf s_{t+h}$ induced by proposed action sequences $\mathbf a_t, \mathbf a_{t+1}, ..., \mathbf a_{t+h}$ such that $\ns \sim p(\cdot|\s,\action)$. If actions are conditioned on the previous state to describe a policy $\pi(\action|\s)$, then planning becomes learning a policy $\pi$ to maximize expected reward over the predicted state-action sequence. A limitation of this approach is that modeling errors are compounded at each time step, resulting in sub-optimal policies when the learning procedure overfits to the imperfect dynamics model. Alternatively, we use \emph{model predictive control (MPC)} to find the action trajectory $\mathbf a_{t:t+h}$ that optimizes $\sum_t^{t+h} \mathbb{E}_{\s,\action \sim p(\mathbf{s_{t}},\mathbf{a}_{t})}[r(\s, \action)]$ at run-time \cite{Camacho}. At each time step, the MPC controller plans into the future, finding a good trajectory over the planning horizon $h$ but applying only the first action from the plan, and re-plans again at the next step. Because of this, MPC is better able to tolerate model bias and unexpected conditions.

Algorithm \ref{al:mpc} describes a learning procedure that uses the cross-entropy method (CEM) as the optimizer for an MPC controller \cite{cem_tutorial}. On each iteration, CEM samples 500 proposed action sequences $\mathbf a_{t:t+h}$ from $h$ independent multivariate normal distributions $\mathcal N (\mathbf a_t|\mu_t,\Sigma_t)$, one for each time step in the planning horizon (line \ref{al:cem-sample}), and calculates the expected reward for each sequence. The top 10\% performing of these are used to update the proposal distribution mean and covariance. However, evaluating the expected reward exactly is intractable, so we use a particle-based approach called trajectory sampling (TS) from \cite{Chua2018} to propagate the approximate next state distributions. For each particle, the latent variable $\mathbf{e}_k \sim q_phi(\cdot|\mathcal{D}_t$ is sampled from approximate posterior so that planning can also account for uncertainty about its environment.

On a new environment, we skip training line \ref{al:train} to keep the dynamics model $f_{\theta}$ fixed. Our task then is to iterate between acting at step $t$ and inferring $p(\env | \mathcal{D}_{t})$ in order to align the expected dynamics with the current system the agent is acting in as it changes.

\vspace{-0.1in}

\begin{algorithm}[!ht]
\caption{Learning and control with Model Predictive Control}
\label{al:mpc}
\begin{algorithmic}[1]
\State Initialize data $\mathcal{D}$ with random policy
\For{Episode m = 1 to M}
    \State Sample an environment indexed by $k$
    \State If learning, train a dynamics model $f_\theta$ with
    $\mathcal{D}$ using Eq. \ref{eq:objective}
    \label{al:train}
    \State Initialize starting state $\mathbf s_0$ and episode history $\mathcal{D}_k = \varnothing$
    \For{Time t = 0 to TaskHorizon}
        \Comment{MPC loop}
        \For{Iteration i = 0 to MaxIter}
            \Comment{CEM loop}
            \State Sample actions $\mathbf{a}_{t:t+h} \sim \mathrm{CEM}(\cdot)$
            \label{al:cem-sample}
            \State Sample latent $\env \sim q_\phi(\env)$ for each particle state particle $s_p$
            \State Propagate next state predictions $s_{t+1,p} \gets s_{t+1,p}$ using $f_\theta$ and TS-$\infty$
            \Comment{See \cite{Chua2018}}
            \State Evaluate expected reward $\sum_{\tau=t}^{t+h} \nicefrac{1}{P}\sum_{p=1}^P r(\mathbf{s}_{\tau,p}, \mathbf{a}_\tau)$
            \State Update $\mathrm{CEM}(\cdot)$ distribution
            \EndFor
        \State Execute first action $\action$ determined by final $\mathrm{CEM}(\cdot)$ distribution
        \State Record outcome $\mathcal{D}_t \gets  \{(\s,\action), \ns\} $
        \State Record outcome $\mathcal{D}_k \gets \mathcal{D}_k \cup \mathcal{D}_t $
        \State Update approximate posterior $q_\phi(\env|\mathcal{D}_t)$ using Eq. \ref{eq:objective_online}
    \EndFor
    \State Update data $\mathcal{D} \gets \mathcal{D} \cup \mathcal{D}_k$
\EndFor
\end{algorithmic}
\end{algorithm}

\section{Experiments}

To demonstrate learning and performance across environments, we show preliminary results in the HalfCheetah Mujoco simulator on proof-of-concept tasks that vary the direction of gravity $\theta_g$ where $0^\circ$ is vertical and positive/negative values tilt forwards/backwards (like walking up or down a hill.) Otherwise environments were set up as in \cite{Chua2018}. We split these environments into training and test environments, $\theta_g^\mathrm{train} \in \{-12^\circ, -6^\circ, 0^\circ, 6^\circ, 12^\circ\}$ and $\theta_g^\mathrm{test} \in \{-15^\circ, -9^\circ, -3^\circ, 3^\circ, 9^\circ, 15^\circ\}$, such that four test environments are interpolated and the two end environments are extrapolated relative to the training distribution. (See Sec.~\ref{app:fewer-training} for results using only 2 training environments). To test continual learning, we test models trained in static environments on a dynamic environment where $\theta_g$ changes from $-6^\circ$ to $6^\circ$ halfway through the episode.
For each environment, we collect an episode rollout via MPC given the current dynamics model. (The first episode is generated via random actions.) Then the ensemble is trained incrementally for 10 iterations on data seen so far, and the process is repeated. At test time, neural network parameters are fixed but the variational parameters are updated online after every time step (using only the last transition) with 60 iterations and $5\times$ larger learning rate. All experiments are repeated with 5 random seeds.

\begin{figure}
    \begin{subfigure}[t]{2.1in}
    \includegraphics[width=2.1in]{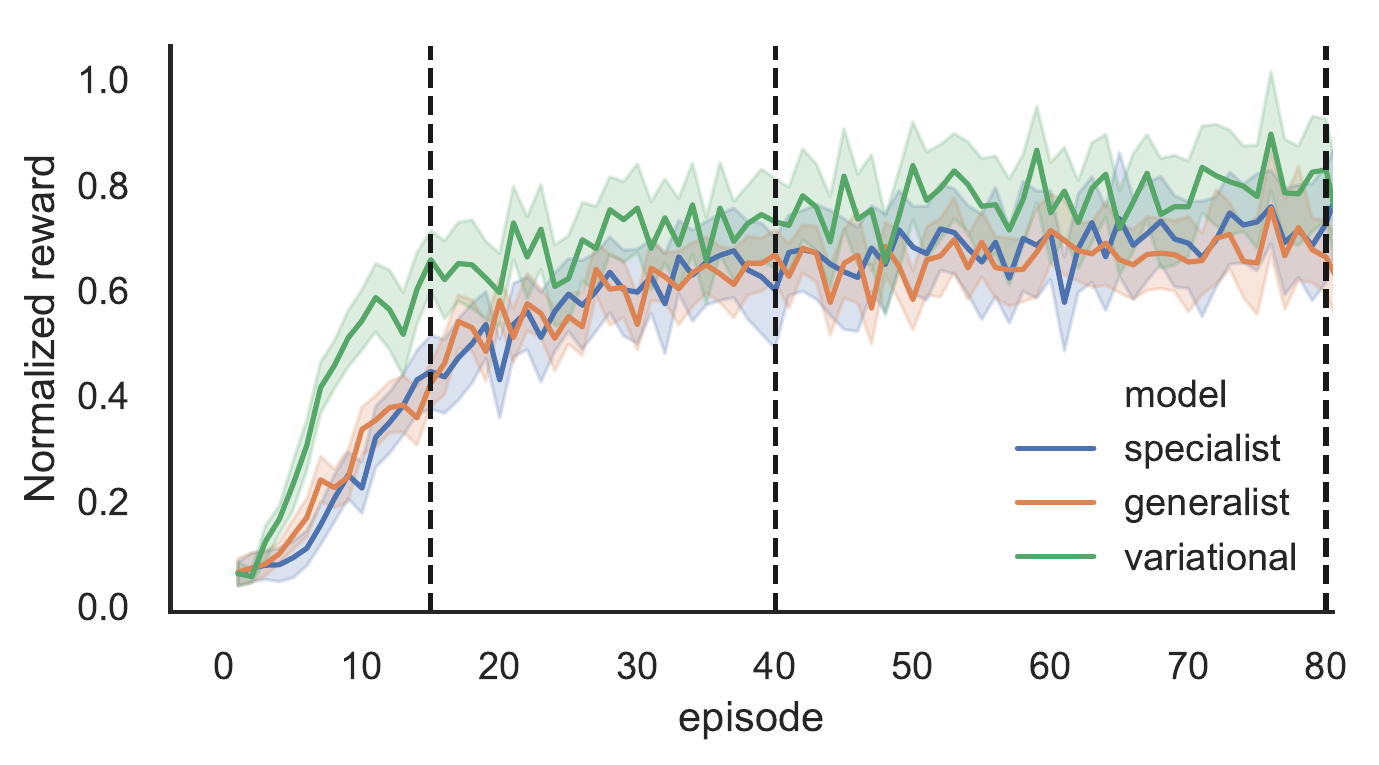}
    \caption{\small Normalized episode reward averaged across training environments.}
    \label{fig:normalized-learning-curve}
    \end{subfigure}
    \begin{subfigure}[t]{3.3in}
    \includegraphics[width=3.3in]{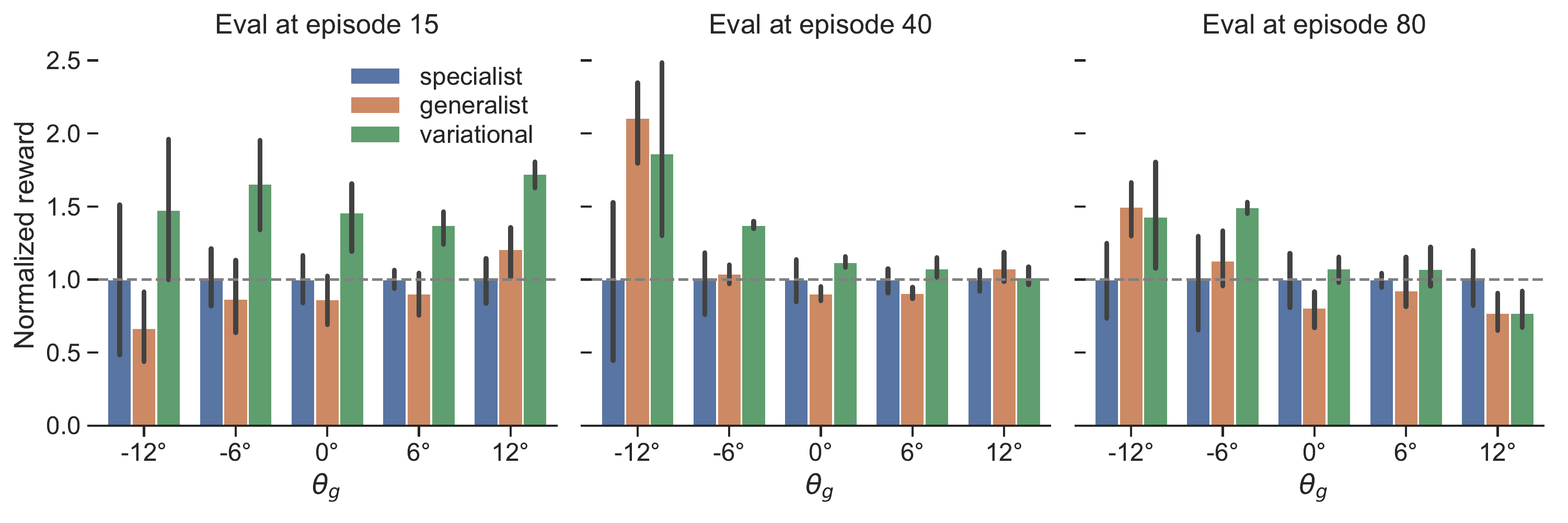}
    \caption{\small Normalized episode reward per training environment.}
    \label{fig:normalized-reward}
    \end{subfigure}
    \caption{\small Comparing normalized episode reward on training environments. \textbf{(a)} For each environment $k$, calculate the best mean reward (averaged over 5 different seeds) across all models $R_k$. The minimum reward $B_k$ is the mean reward from 50 rollouts of a random policy.  Then, the observed rewards at episode $m$ are rescaled by the maximum and minimum mean reward, i.e. $\hat r = \nicefrac{r-B_k}{R_k-B_k}$. The shaded regions and error bars are 95\% CI from 500 bootstraps. \textbf{(b)} Rewards are normalized to the specialist mean (=1) and random policy (=0) per environment. The dashed vertical lines indicate checkpoints 15, 40, and 80 where the models are also evaluated on novel test environments (see Fig.~\ref{fig:rewards}, lower panel).}
    \vspace{-4mm}
\end{figure}

We compare our method against two baselines, a specialist and generalist agent. A specialist is an ensemble dynamics model (from \cite{Chua2018}) trained separately for each environment (no auxiliary variables or inference) and executed via MPC, achieving impressive performance in only dozens of episodes. This single-environment agent provides a strong performance baseline, but as we will see, can be beat because of positive transfer during training. The generalist is functionally the same as the specialist, but is trained on all data collected from the environments in $\theta_g^\mathrm{train}$ but also \emph{without auxiliary latent variables}. Thus the generalist demonstrates the ability of the model-based ensemble to generalize without our approach. Our latent model uses variational inference to jointly learn both a dynamics model and a latent representation $e_k$ of environment parameters, and so is expected to compare favourably against the generalist.

\begin{wrapfigure}[21]{r}{0.65\textwidth}
  \begin{subfigure}[b]{.65\textwidth}
    \includegraphics[width=\textwidth]{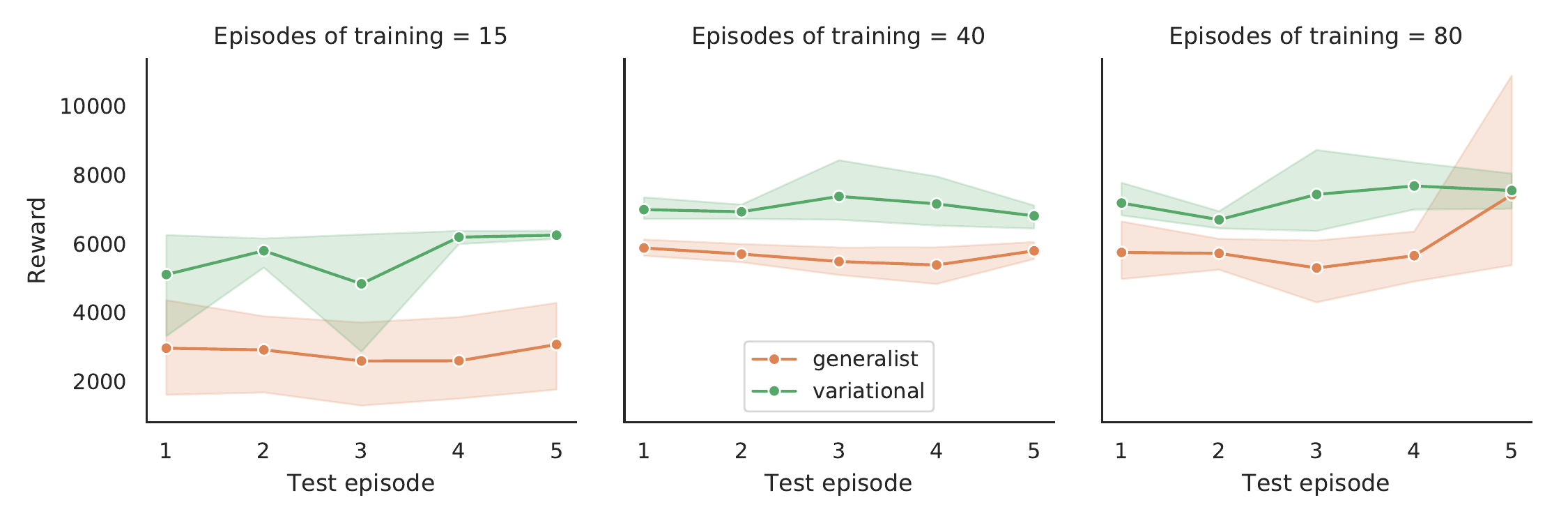}
  \end{subfigure}

  \begin{subfigure}[b]{.65\textwidth}
    \includegraphics[width=\textwidth]{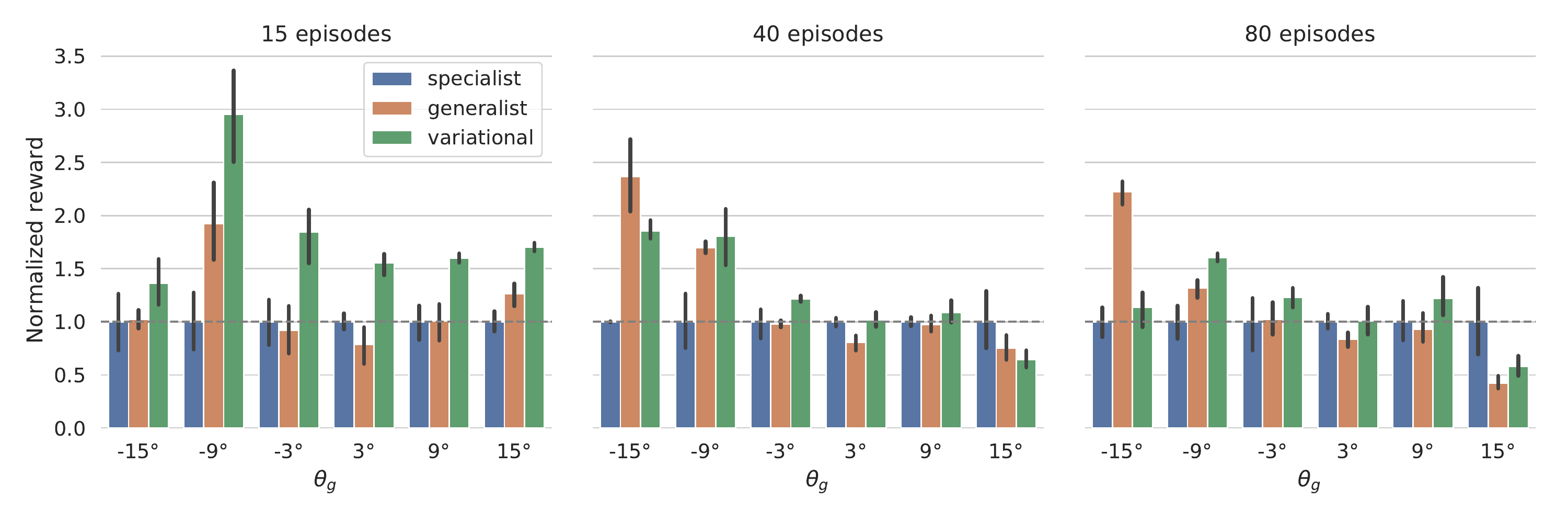}
  \end{subfigure}
  \caption{\small \textbf{Top:} Model performance on  dynamically changing environment. $\theta_g$ changes from $-6^\circ$ to $6^\circ$ after 500 time steps in each episode. \textbf{Bottom:} Final normalized scores obtained on the novel environments after a few episodes of continuous learning.}
  \label{fig:rewards}
\end{wrapfigure}

Fig.~\ref{fig:normalized-learning-curve} shows the training performance as a function of the number of episodes seen per environment. Overall, the variational approach learns faster than both baselines, doing at least as well as an agent specialized for each environment. However the performance difference closes with more training data (see Fig.~\ref{fig:normalized-reward} for a breakdown), as expert baselines can sometimes learn to model the system arbitrarily well given enough data.
We explore 5 episodes of online learning on an environment with changing dynamics, showing that the variational approach outperforms the generalist (Fig.~\ref{fig:rewards}; Top). On the static (Bottom) test environment, the variational approach is superior in the low-data regime, demonstrating positive transfer without sacrificing performance overall. Yet its ability to extrapolate to the most extreme test angles does not exceed that of the generalist significantly. Either the ensemble-based generalist is a stronger baseline than anticipated for this task or further algorithmic improvements remain to be explored.

\begin{figure}
    \begin{subfigure}[t]{0.6\textwidth}
    \includegraphics[width=\textwidth]{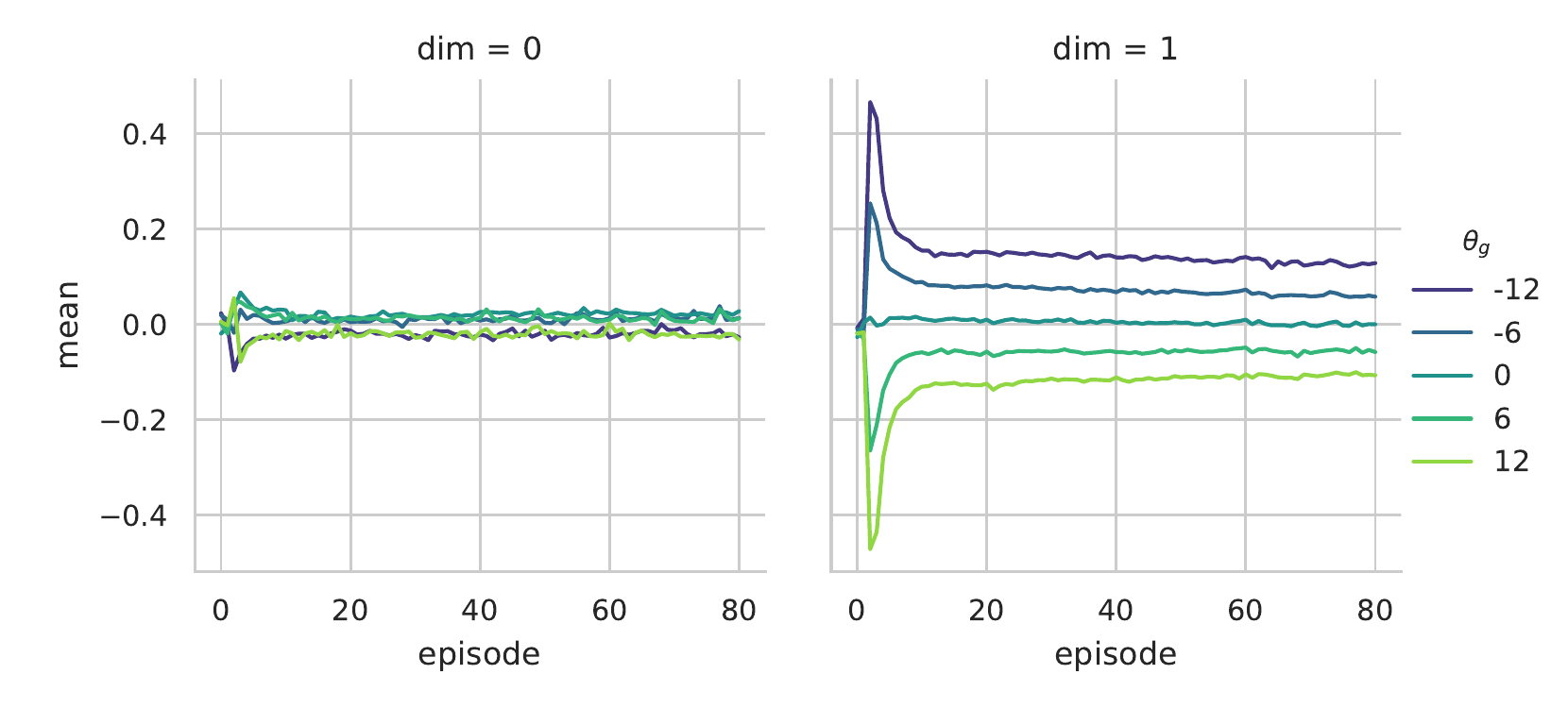}
    \end{subfigure}
    \begin{subfigure}[t]{0.3\textwidth}
    \raisebox{3mm}{\includegraphics[width=\textwidth]{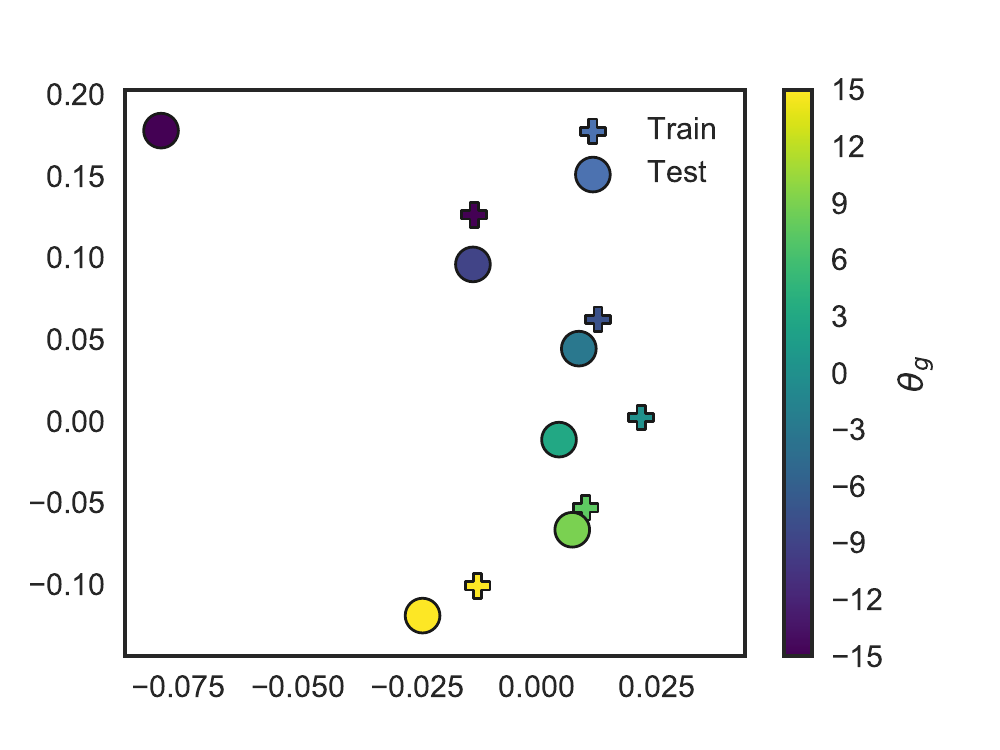}}
    \vfil
    \end{subfigure}
    \caption{\small \textbf{Learned latent variable.} \textbf{Left and Center:} Means of the two dimensions of the latent variable during training. Dimension two is in order of $\theta_g$. \textbf{Right:} Mean of 2D inferred latent variables from training and test environments.}
    \label{fig:latents}
    \vspace{-2mm}
\end{figure}

As seen in Fig.~\ref{fig:latents}, we observe that the unsupervised latent embeddings of dynamics capture the relative order of the hidden parameter $\theta_g$ on training and test environments is captured accurately, while the $\mathbf e_k$ of the steepest novel test environments are inferred to lie outside those of the training environments.

\section{Discussion}

A long-standing aim in reinforcement learning is the search for robust learning algorithms that learn efficiently with less data and produce adaptable agents that generalize to many situations. We have shown that (ensemble) dynamics models with auxiliary latent variables can be used to learn quickly across environments with varying physical parameters, and also that they allow robust control on novel environments. This suggests that the latent variable approximately captures relevant degrees of freedom of the true dynamics of the environment and utilizes an explicit belief about the state of its environment beneficially. Our method suggests an alternate approach to popular meta-learning approaches (such as MAML \cite{finn2017}) to learning adaptive models across environment dynamics, and can readily incorporate other information available to a robotic agent, e.g. other perceptions or informative priors. Online inference relaxes constraints on environment stationarity without resorting to, e.g., sample-inefficient recurrent neural network policies.
This combination of modeling elements affords our agents strong performance in terms of data efficiency and generalization on high-dimension continuous control task. Our results suggest future work on harder and more varied environments to further test the performance benefit of auxiliary latent variables.

\newpage
\bibliographystyle{plain}
\bibliography{pmbrl}

\begin{thebibliography}{10}

\bibitem{Camacho}
E.~F. Camacho and C.~Bordons.
\newblock {\em Model Predictive Control}.
\newblock Springer Science \& Business Media, 2013.

\bibitem{Chua2018}
Kurtland Chua, Roberto Calandra, Rowan McAllister, and Sergey Levine.
\newblock {Deep Reinforcement Learning in a Handful of Trials using
  Probabilistic Dynamics Models}.
\newblock {\em NeurIPS}, arXiv:1805.12114v1, 2018.

\bibitem{Clavera2018}
Ignasi Clavera, Anusha Nagabandi, Ronald~S. Fearing, Pieter Abbeel, Sergey
  Levine, and Chelsea Finn.
\newblock {Learning to Adapt: Meta-Learning for Model-Based Control}.
\newblock {\em CoRR}, abs/1803.11347, 2018.

\bibitem{cem_tutorial}
Pieter-Tjerk De~Boer, Dirk~P Kroese, Shie Mannor, and Reuven~Y Rubinstein.
\newblock A tutorial on the cross-entropy method.
\newblock {\em Annals of operations research}, 134(1):19--67, 2005.

\bibitem{Deisenroth2011}
Marc~P Deisenroth and Carl~E. Rasmussen.
\newblock {PILCO: A Model-Based and Data-Efficient Approach to Policy Search}.
\newblock {\em Proceedings of the International Conference on Machine
  Learning}, 2011.

\bibitem{Deisenroth2011a}
Marc~Peter Deisenroth.
\newblock {A Survey on Policy Search for Robotics}.
\newblock {\em Foundations and Trends in Robotics}, 2011.

\bibitem{doshi2016hidden}
Finale Doshi-Velez and George Konidaris.
\newblock Hidden parameter markov decision processes: A semiparametric
  regression approach for discovering latent task parametrizations.
\newblock In {\em IJCAI: proceedings of the conference}, volume 2016, page
  1432. NIH Public Access, 2016.

\bibitem{finn2017}
Chelsea Finn, Pieter Abbeel, and Sergey Levine.
\newblock Model-agnostic meta-learning for fast adaptation of deep networks.
\newblock {\em Proceedings of the 34th International Conference on Machine
  Learning}, abs/1703.03400, 2017.

\bibitem{Gal2016}
Yarin Gal, Rowan~Thomas Mcallister, and Carl~Edward Rasmussen.
\newblock {Improving PILCO with Bayesian Neural Network Dynamics Models}.
\newblock In {\em Data-Efficient Machine Learning Workshop, ICML}, 2016.

\bibitem{haruno2001mosaic}
Masahiko Haruno, Daniel~M Wolpert, and Mitsuo Kawato.
\newblock Mosaic model for sensorimotor learning and control.
\newblock {\em Neural computation}, 13(10):2201--2220, 2001.

\bibitem{hausman2018}
Karol Hausman, Jost~Tobias Springenberg, Ziyu Wang, Nicolas Heess, and Martin
  Riedmiller.
\newblock Learning an embedding space for transferable robot skills.
\newblock {\em ICRL}, 2018.

\bibitem{killian2017robust}
Taylor~W Killian, Samuel Daulton, George Konidaris, and Finale Doshi-Velez.
\newblock Robust and efficient transfer learning with hidden parameter markov
  decision processes.
\newblock In {\em Advances in Neural Information Processing Systems}, pages
  6250--6261, 2017.

\bibitem{balaji2017}
Balaji Lakshminarayanan, Alexander Pritzel, and Charles Blundell.
\newblock {Simple and Scalable Predictive Uncertainty Estimation using Deep
  Ensembles}.
\newblock {\em NIPS}, arXiv:1612.01474, 2017.

\bibitem{Mnih2015}
Volodymyr Mnih, Koray Kavukcuoglu, David Silver, Andrei~A. Rusu, Joel Veness,
  Marc~G. Bellemare, Alex Graves, Martin Riedmiller, Andreas~K. Fidjeland,
  Georg Ostrovski, Stig Petersen, Charles Beattie, Amir Sadik, Ioannis
  Antonoglou, Helen King, Dharshan Kumaran, Daan Wierstra, Shane Legg, and
  Demis Hassabis.
\newblock {Human-level control through deep reinforcement learning}.
\newblock {\em Nature}, 2015.

\bibitem{nagabandi2018}
Anusha Nagabandi, Gregory Kahn, Ronald~S Fearing, and Sergey Levine.
\newblock Neural network dynamics for model-based deep reinforcement learning
  with model-free fine-tuning.
\newblock In {\em 2018 IEEE International Conference on Robotics and Automation
  (ICRA)}, pages 7559--7566. IEEE, 2018.

\bibitem{yao2018}
Jiayu Yao, Taylor Killian, George Konidaris, and Finale Doshi-Velez.
\newblock Direct policy transfer via hidden parameter markov decision
  processes.
\newblock {\em LLARLA Workshop, FAIM 2018}, 2018.

\end{thebibliography}

\newpage
\appendix

\section{Learning environment dynamics}
\label{sec:learning-dynamics}

We define the transition model of an environment by $T_{\eta}(\ns|\s,\action)$ parameterized by $\eta$, which includes physical constants like gravity, friction, and dampening, or properties of an agent like actuator gain or noise \cite{doshi2016hidden, killian2017robust}. In order to perform model-based control for an agent acting in such an environment, one requires knowledge of the transition dynamics, which are composed of the dynamic mechanisms and the constants $\eta$.

When the quantities $\eta$ and the underlying mechanisms that govern environment dynamics are unknown, one can resort to learning a model of these dynamics $f_\theta(\ns|\s,\action)=p(\ns|\s,\action)$ given data observed from the environment $\mathcal{D}=\{(
\s,\action),\ns\}_{t=1}^N$.

Because environment dynamics can be stochastic, one can use a generative model of transition dynamics. Since these are continuous quantities, they can be modeled with a Gaussian likelihood parameterized by mean $\mu_\theta$ and covariance $\Sigma_\theta$ by a neural network $f_\theta$ with parameters $\theta$:
\begin{equation}
\begin{split}
p(\ns|\s,\action) & = p(\ns|\s,\action;\theta)\\
& = \mathcal{N}\left(\mathbf{\mu}_\theta(\s, \action), \Sigma_\theta(\s, \action)\right)
\end{split}
\label{eq:likelihood}
\end{equation}

Here as elsewhere, instead of $\ns$, a neural network predicts the change in the states $\Delta_s = \ns - \s$ given the state and action $p(\Delta_s | \s, \action)=f_\theta$.

\subsection{Ensembles of networks}
\label{app:ensembles}

In order to be robust to model mis-specification and handle the small data setting, one can model uncertainty about parameters $\theta$ and marginalize to obtain
\begin{equation}
p(\ns|\s,\action) = \int p(\ns|\s,\action,\theta) p(\theta)  d\theta \,.
\end{equation}

However, Bayesian neural networks can be computationally expensive to train. A practical way to approximate drawing samples from the posterior $p(\theta|\mathcal{D})$ is through ensembles of predictors each trained on different bootstraps or shuffles of the training data \cite{balaji2017}:

\begin{equation}
    p(\ns|\s,\action) = \frac{1}{|E|}\sum_{\theta \in E} p_\theta(\ns|\s,\action) \,.
    \label{eq:ensemble}
\end{equation}

In this work, each member of the ensemble is trained on distinct shuffles of the same data, and are reshuffled at every epoch as suggested in \cite{balaji2017}. In contrast, Chua et al. \cite{Chua2018} use bootstrap samples.
% The predictive distribution after having observed data $\mathcal{D}$ utilizes the posterior distribution $p(\theta|\mathcal{D})$ is
% \begin{equation}
% p(s_{t+1}|s_t,a_t, \mathcal{D}) = \int p( s_{t+1} | s_t, a_t ,\theta) p(\theta|\mathcal{D}) d\theta \,.
% \end{equation}

\subsection{Learning curves}

\begin{figure}[!hb]
    \includegraphics[width=\textwidth]{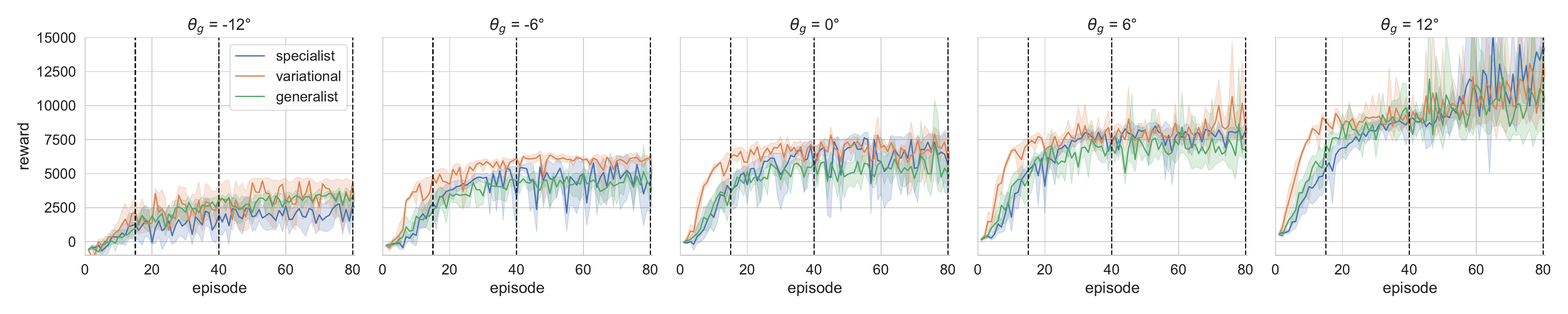}
    \caption{Episode reward per environment as a function of episodes seen on that environment. This implies that the generalist and variational models see 5X as much data as the specialist in total. The dashed vertical lines indicate checkpoints 15, 40, and 80 where the models are evaluated on novel test environments.}
    \label{fig:learning-curve}
\end{figure}

\section{Implementation}

The ELBO for a single environment $\mathcal D_k$ with $T$ timepoints is

\begin{eqnarray}
\label{eq:step1}
\log p_\theta(\mathcal D_k | e_k) &\geq \mathbb{E}_{\env \sim q_\phi(\env)}\left[ \sum_{t=0}^{T} \log p_\theta(\ns|\s,\action,\env) \right] - \mathrm{KL}\big(q_\phi(\env) || p(\env) \big) \\
\label{eq:step2}
&\geq  \mathbb{E}_{\env \sim q_\phi(\env)}
\left[\sum_{t=0}^{T} \log p_\theta(\ns | \s,\action,\env)\right] - \mathrm{KL}\big(q_\phi(\env) || p(\env) \big).
\end{eqnarray}

For clarity, we define $\log p_\theta(\mathcal D_k | e_k) = \sum_{t=0}^{T} \log p_\theta(\ns | \s,\action,\env)$. Given ensemble $E$ of networks and Eq. \ref{eq:ensemble}, the complete data likelihood is an expectation over ensembles. Note that because each ensemble member is trained independently on different data, the expectation is \emph{outside} the $\log$. Hence,

\begin{equation}
    \log p(\mathcal D_k) = \mathbb E_{\theta \sim E} [\log p_\theta(\mathcal D_k)] \geq \mathbb{E}_{\env \sim q_\phi(\env), \theta \sim E} \left[ \log p_\theta(\mathcal D_k|\env) \right] - \mathrm{KL}\big(q_\phi(\env) || p(\env) \big) \,.
\end{equation}

\section{Generalizing from fewer environments}
\label{app:fewer-training}

As in multi-task and meta-learning scenarios, the number of training tasks to sample is an important hyperparameter. Below, both the variational method and the generalist baseline are trained on only two environments $\theta_g = \pm6^\circ$ and evaluated after 100 episodes. In contrast to 5 environment training in Fig.~\ref{fig:normalized-learning-curve}, the advantage of the variational method grows larger with more training, and remains competitive with the specialist on novel environments except for the steepest forward sloping environment (where the specialist gets very high reward.)

\begin{figure}[!ht]

\begin{subfigure}{\textwidth}
    \centering
    \includegraphics[width=4in]{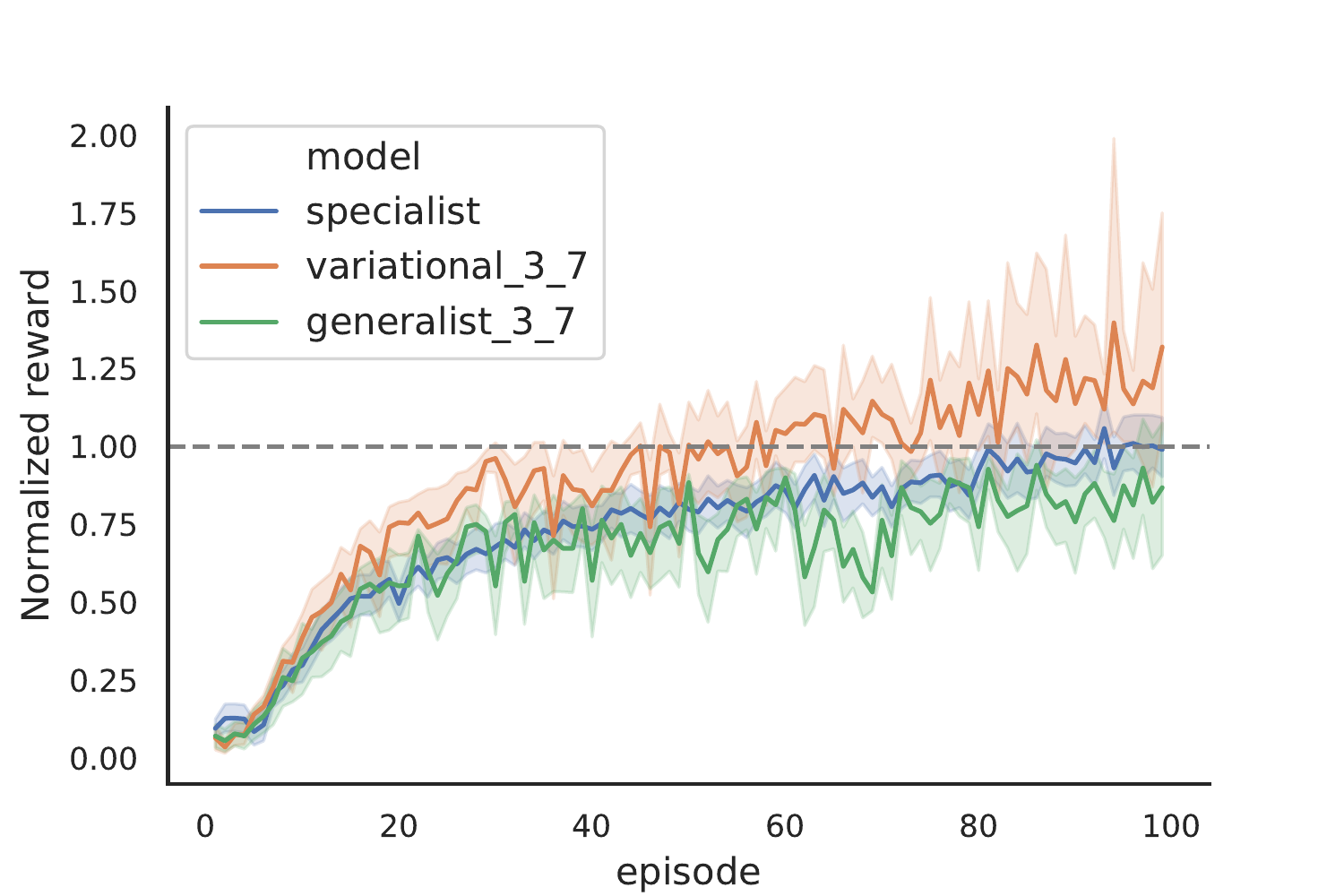}
    \caption{Learning curve comparing variational method against generalist. In contrast to Fig.~\ref{fig:normalized-learning-curve}, both panels in this figure are normalized such that 1 equals the mean reward of the specialist after 100 episodes.}
\end{subfigure}

\begin{subfigure}{\textwidth}
    \centering
    \includegraphics[width=4in]{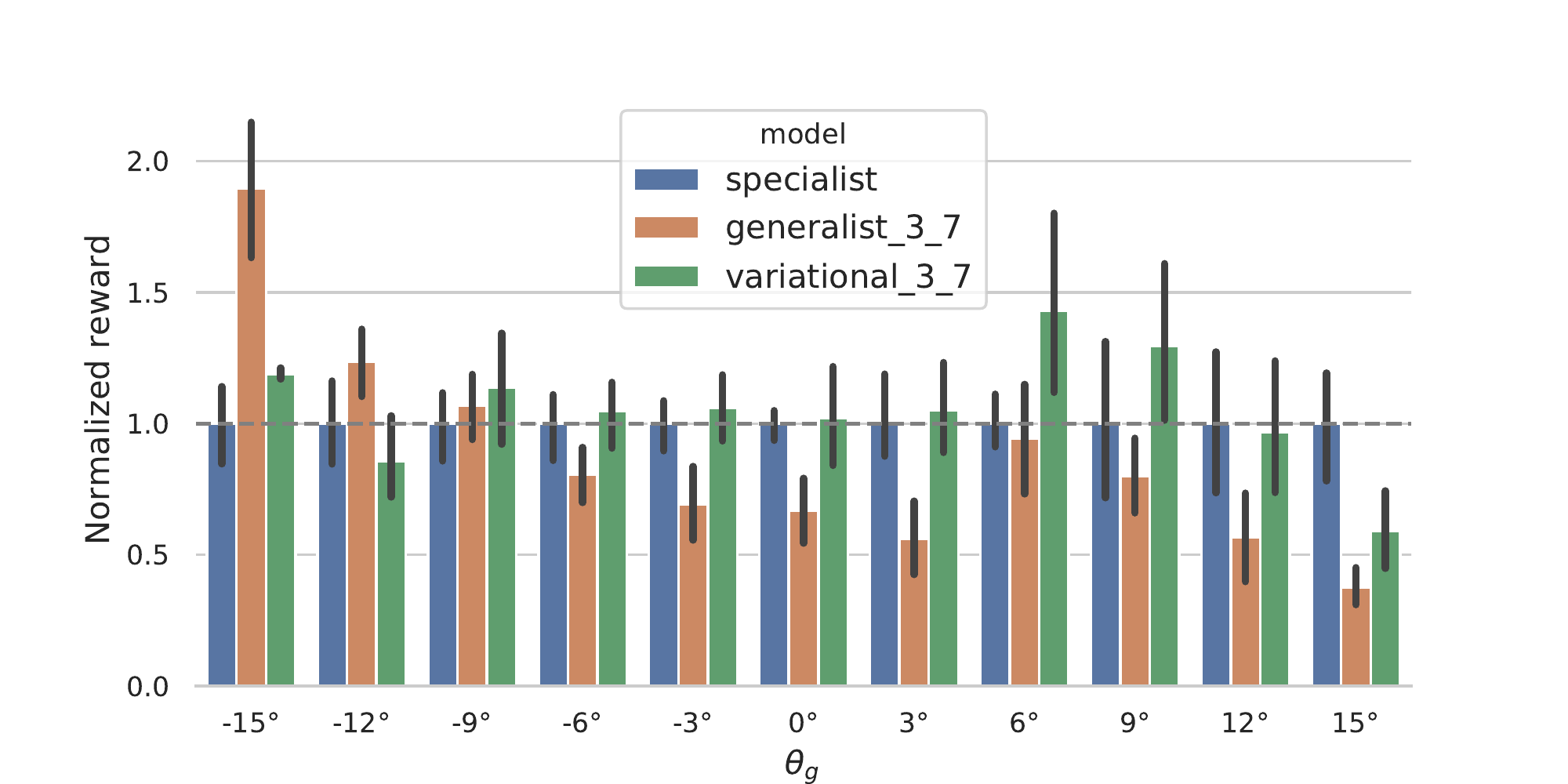}
    \caption{Normalized reward on training and test environments after 100 episodes.}
\end{subfigure}
\end{figure}

\end{document}